\documentclass{article}

\usepackage{arxiv}

\usepackage[utf8]{inputenc} 
\usepackage[T1]{fontenc}    
\usepackage{hyperref}       
\usepackage{url}            
\usepackage{booktabs}       
\usepackage{amsfonts}       
\usepackage{nicefrac}       
\usepackage{microtype}      
\usepackage{lipsum}
\usepackage{graphicx}
\graphicspath{ {./images/} }

\title{ClamNet: Using contrastive learning with variable depth Unets for medical image segmentation}

\date{June 9, 2022}	

\author{ \href{https://orcid.org/0000-0002-2725-5367}{\includegraphics[scale=0.06]{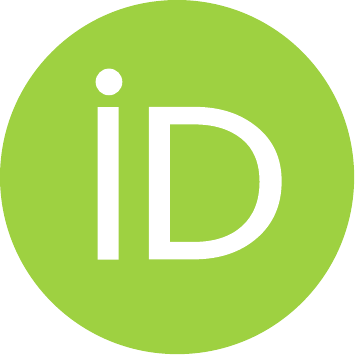}\hspace{1mm}Samayan Bhattacharya} \\
	Department of Computer Science and Engineering\\
	Jadavpur University\\
	Kolkata, India \\
	\texttt{samayan.bhattacharya@gmail.com} \\
	\And
	\href{https://orcid.org/0000-0001-7480-427X}{\includegraphics[scale=0.06]{orcid.pdf}\hspace{1mm}Sk Shahnawaz$^*$} \\
	Department of Computer Science and Engineering\\
	Jadavpur University\\
	Kolkata, India \\
	\texttt{skshahnawaz2909@gmail.com} \\
	\And
	\href{https://orcid.org/0000-0002-9567-1346}{\includegraphics[scale=0.06]{orcid.pdf}\hspace{1mm}Avigyan Bhattacharya$^*$} \\
	Department of Computer Science and Engineering\\
	Jadavpur University\\
	Kolkata, India \\
	\texttt{avigyanbhattacharya123@gmail.com} \\
}

\begin{document}
\maketitle
\def\thefootnote{*}\footnotetext{These authors contributed equally to this work}
\begin{abstract}
Unets have become the standard method for semantic segmentation of medical images, along with fully convolutional networks (FCN). Unet++ was introduced as a variant of Unet, in order to solve some of the problems facing Unet and FCNs. Unet++ provided networks with an ensemble of variable depth Unets, hence eliminating the need for professionals estimating the best suitable depth for a task. While Unet and all its variants, including Unet++ aimed at providing networks that were able to train well without requiring large quantities of annotated data, none of them attempted to eliminate the need for pixel-wise annotated data altogether. Obtaining such data for each disease to be diagnosed comes at a high cost. Hence such data is scarce. In this paper we use contrastive learning to train Unet++ for semantic segmentation of medical images using medical images from various sources including magnetic resonance imaging (MRI) and computed tomography (CT), without the need for pixel-wise annotations. Here we describe the architecture of the proposed model and the training method used. This is still a work in progress and so we abstain from including results in this paper. The results and the trained model would be made available upon publication or in subsequent versions of this paper on \emph{arxiv}.
\end{abstract}


\section{Introduction}

\begin{figure}[htp]
    \centering
    \includegraphics[width=15cm]{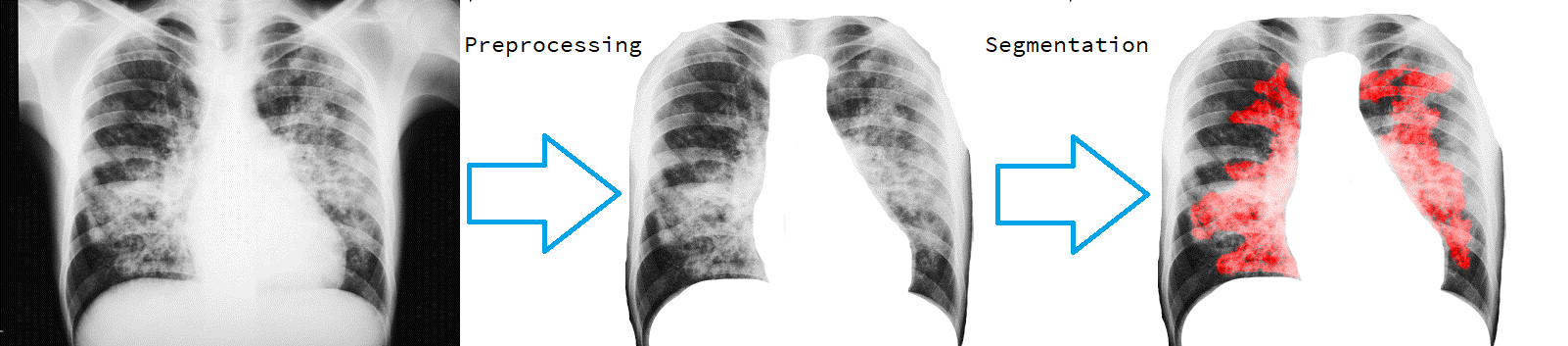}
    \caption{Different stages of image transformation by the proposed method from input chest X-ray image (on the left) to output segmentation (on the right)}
    \label{fig:galaxy}
\end{figure}

Image segmentation is highly important for medical images as it allows human professionals to deal with the more challenging patients, leaving the ordinary patients to the automatic system. As opposed to classification, segmentation provides some insight into how the system is analyzing the image and hence is more reliable for applications at scale.

Encoder-decoder networks are widely used for semantic segmentation of images \cite{1,2,3,4,5,6}. Unet \cite{100} introduced skip connections between the encoder and the decoder. This improved performance of the network by allowing the convolutional layers in the decoder to access the fine-grained feature maps from the shallow layers of the encoder. However, the appropriate depth of the network for a particular application had to be determined empirically or heuristically, leading to a waste of resources and suboptimal results. The standard way to address this issue was to train networks of different depths and take an ensemble of these at inference time \cite{14,15,16}. However, this practice was inefficient while deploying at scale \cite{17,18}. Another issue was that the feature maps had to be of the same size to be combined. However, there was no theoretical guarantee that this was the most efficient combination.

Unet++ \cite{100} was introduced with a built-in ensemble of Unets of different depths and was able to combine feature maps of different sizes. Hence it was able to perform better than classical Unets.It achieved these by including additional convolutional units in between the encoder and decoder parts of the standard Unet architecture.This also allowed the model to be pruned to improve efficiency without adversely affecting performance. However, it still relied heavily on pixel-wise annotated data to train. Such annotations are tediously obtained at the cost of a lot of man-hours and financial expenditure. Hence, large pixel-wise annotated datasets of medical images are hard to come by. Datasets with image specific labels are much more easily available. However, these are used for classification rather than semantic segmentation of images. Classification networks work as a black box and the only way to gain insights into what part of the image the network focussed on, while classifying it as belonging to a particular class, can only be obtained by using tools like gradcam \cite{101}. Such tools analyze the gradient map of a particular convolutional layer and indicates the region of interest. For medical images, the region of interest may or may not be restricted to the abnormality we are looking for in an image to diagnose an underlying medical condition. For example, for diagnosing pneumonia using CT images, a patch might occur in different parts of the lung and hence the whole lung would be the region of interest for the network. This does not indicate if there is a patch or where it is.

Contrastive learning \cite{1-22} is commonly used for self-supervised learning. It works by making sure that the pixels with same label produce the same output in two identical models, while the pixels with different labels produce different outputs. In this paper, we use contrastive learning in an unsupervised setting.

In this paper we propose to use contrastive learning to train a pair of unet++ models using labelled images without pixel-wise annotations. For this purpose, we use a probabilistic loss term. We use our model on different types of medical images, including X-ray, magnetic resonance imaging (MRI) and computed tomography (CT). We also test our model for the diagnosis of different medical conditions, including pneumonia, carcinoma, trauma, etc. Preliminary results are promising and we would present a comprehensive study upon publication of our work or in subsequent versions of this paper on \emph{arxiv}.

\section{Related works}

\subsection{Self-supervised contrastive learning}

This approach works by contrasting the outputs of two identical models for positive pairs (pairs of inputs belonging t the same class) and negative pairs. In recent years several approaches have used contrastive loss \cite{1-22} for learning representations of visual information \cite{1-9, 1-10, 1-14, 1-24, 1-35, 1-57, 1-60}. These approaches use data augmentation techniques to generate positive pairs and contrast them with other instances. Some approaches use hard negative mining techniques \cite{1-29,1-45}. Some approaches used data banks to store representations as performance is seen to improve with increase in number of negative pairs \cite{1-24, 1-51, 1-57}. Some approaches have extended its use to semantic segmentation of images by classification of individual pixels\cite{1-4, 1-43, 1-55, 1-58, 1-59} and by using large datasets of pixel-wise annotate images. 

\subsection{Semantic segmentation}

Convolutional Neural Networks (CNNs) have been used for segmentation of images for a long time \cite{1-17,1-38}. Such models have undergone incremental improvements by using more convolutional layers in deeper networks \cite{1-6, 1-7, 1-8, 1-21, 1-53,1-61, 1-62, 1-63, 1-64}. An alternative technique is pretraining the model on a large dataset, for example the ImageNet classification dataset or weak supervision techniques like unlabeled images \cite{5, 19, 39, 40, 48, 67}, bounding boxes \cite{12, 31, 42, 47}, image-level labels \cite{1, 26, 34, 42} or points \cite{2} and scribbles \cite{36, 50}. 

Another approach learns pixel relations by using region-based loss function \cite{1-30,1-65}. A region Mutual Information (MI) loss is used to maximize the MI between patch label distributions \cite{1-65}. A pairwise affinity loss based on
KL divergence between the probabilities, predicted for belonging to each class, is used by \cite{1-30}. Some works by \cite{1-3, 1-18, 1-23, 1-33} performed instance and semantic segmentation by using metric learning based on similarity and dissimilarity pairs. The work by \cite{1-28} trained the feature extractor of a semantic segmentation model by maximizing log likelihood of an extracted feature under several vMF distribution models. At the time of inference, they used K-means clustering to segment the pixel features and used K nearest neighbours to get labels from available segments.

\subsection{Unet++}
Since Unet++ was first introduced by \cite{51}, it has been used in multiple ways. \cite{52,53,54,55} used it as a baseline, while \cite{56,57,58,59,60,61} proposed models inspired by the Unet++ architecture. It was used for natural image segmentation by \cite{64}, for biomedical image segmentation by \cite{62,63} and for satellite image segmentation \cite{65,66}. It has also been used for contact prediction by \cite{67}.

\section{Proposed model}

\begin{figure}[htp]
    \centering
    \includegraphics[width=14cm]{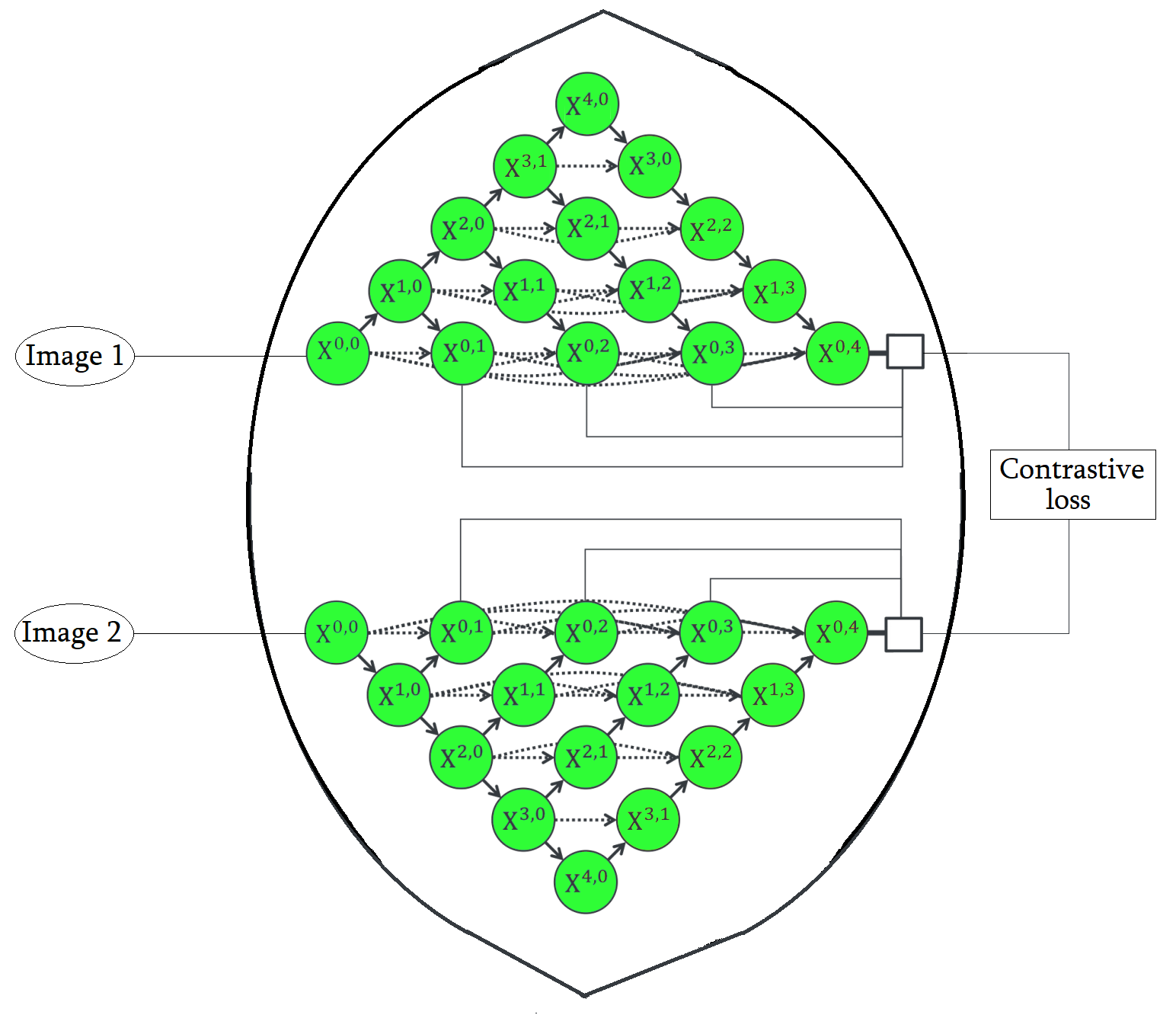}
    \caption{Proposed model architecture. The shape resembles a clam. The name might be altered at a later stage of our work.}
    \label{fig:galaxy}
\end{figure}

\subsection{Unet++ architecture}

We use a classical unet++ architecture with 5 levels of convolutional layers. The input image is of size 256x256. The subsequent feature maps have double the number of channels of the layer above it and length and breadth dimensions are reduced by half, using stride=2 for the second convolution of the two convolutions at each level. The kernel size is higher in lower levels allowing more context information. The intermediate convolutions are taken according to the architecture of the original unet++.

As an alteration of the above approach, we try having two subsequent layers with the same dimensions of feature map. We make this change for randomly selected levels. Initial results indicate that this approach gives better results than a deeper unet++.

\subsection{Preprocessing}

In this study we use labelled images with image-wise annotations instead of pixel-wise annotations. The images are then passed to an image segmentation network that is trained to detect the organ of interest. Segmentation of organs is much easier due to easy availability of data. This is because humans have few organs but each organ has many diseases. After segmentation, the image is cropped such that it contains minimum background. This process is helpful for the probabilistic loss function we describe later in this paper. Each image is resized to 256x256 pixels.

\subsection{Contrastive training}

We take two identical models, as indicated above.These are trained simultaneously on positive pairs and negative pairs of slices of the image. The target is to have both networks produce the same output for positive pairs and different outputs for negative pairs.

Positive pair is a pair of slices belonging to the same class label. We generate a positive pair by (i) blurring the image (such that resultant image has a resolution between 90 to 100 percent of the original resolution) (ii) stretch and compression distortions (such that resultant image has dimensions between 80 to 100 percent of the original resolution), followed by resizing the image to the original size or padding it to the original size. Apart from these, we also generate positive pairs by taking slices from images labelled as normal (negative for the medical condition to be diagnosed).

Negative pairs are pairs of slices that belong to different label classes. These are generated by taking a slice from a positive image and one from a negative image. Since the whole of the positive image does not contain the features of the medical condition to be diagnosed, we use the probabilistic loss function.

\subsection{Probabilistic loss function}

Since we are trying to produce pixel-wise annotated images by training the model on image-wise annotated images, there is no way to be sure which part of the image has markers of the medical condition to be diagnosed. But we increase the probability of the marker being in a given region of the image by segmenting the organ which is supposed to be affected and removing the background. Now, we consider the probability of a marker being present in a given region of the image while calculating the loss.

Mathematically this is given as,

\begin{equation}
	Hybrid\_loss(Y,P)=-\frac{1}{N}.\sum _{c=1}^{C}\sum _{n=1}^{N}(y_{n,c}.log p_{n,c}+ \frac{y_{n,c}.p_{n,c}}{y_{n,c}^2+p_{n,c}^2})
\end{equation}
where $y_{n,c} \in Y$, representing the target labels and $p_{n,c} \in P$ 
representing the predicted probabilities for class c and $n^{th}$ pixel in the batch, where $n \in N$, representing the number of pixels in one batch.

The total loss of the model is the weighted sum of the individual hybrid losses, given by $\sum_{i=1}^{d}(\eta.Hybrid\_loss(Y,P^i))$, where d is the number of slices in the batch and $\eta$ is the weight that is the probability of the slice containing the marker for the medical condition to be diagnosed. If this probability is not known for a dataset, it is treated as a hyperparameter, to be tuned.

\section{Conclusion}

In this paper we proposed a novel method for using contrastive learning, in an unsupervised setting, to train a UNET++ architecture. We are in the process of trying out different methods and testing them on different datasets. This paper is intended to provide an overview of the method. We do not make any claims about its performance in this paper. We plan to provide a comprehensive report of our final model, datasets used to train it and a comparative study of the results obtained for our model versus other commonly used architectures, upon publication of our work or in subsequent versions of this paper on \emph{arxiv}.

\bibliographystyle{unsrt}  


\end{document}